\documentclass[sigconf]{acmart}

\AtBeginDocument{%
  \providecommand\BibTeX{{%
    \normalfont B\kern-0.5em{\scshape i\kern-0.25em b}\kern-0.8em\TeX}}}

\acmDOI{}
\acmISBN{}
\setcopyright{rightsretained}

\makeatletter
\gdef\@copyrightpermission{
\begin{minipage}{0.3\columnwidth}
\href{https://creativecommons.org/licenses/by/4.0/}{\includegraphics[width=0.90\textwidth]{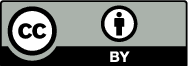}}
\end{minipage}\hfill
\begin{minipage}{0.7\columnwidth}
\href{https://creativecommons.org/licenses/by/4.0/}{This work is licensed under a Creative Commons Attribution International 4.0 License.} \end{minipage} \vspace{5pt} } \makeatother

\copyrightyear{2023}
\acmYear{2023}
\acmConference[WTF 2023]{Working with Trouble and Failures
in conversation between humans and robots}{June 19,
  2023}{Eindhoven, Netherlands}

\usepackage{enumitem}

\begin{document}

\title{Chat Failures and Troubles: Reasons and Solutions}

\author{Manal Helal}
\email{m.helal@herts.ac.uk}
\orcid{1234-5678-9012}
\affiliation{%
  \institution{University of Hertfordshire}
  \department{School of Physics, Engineering and Computer Science}
  \streetaddress{College Lane}
  \city{Hatfield}
  \country{United Kingdom}
  \postcode{AL10 9AB}
}

\author{Patrick Holthaus}
\email{p.holthaus@herts.ac.uk}
\orcid{0000-0001-8450-9362}
\affiliation{%
  \institution{University of Hertfordshire}
  \department{School of Physics, Engineering and Computer Science}
  \streetaddress{College Lane}
  \city{Hatfield}
  \country{United Kingdom}
  \postcode{AL10 9AB}
}

\author{Gabriella Lakatos}
\email{g.lakatos@herts.ac.uk}
\orcid{0000-0003-1436-7324}
\affiliation{%
  \institution{University of Hertfordshire}
  \department{School of Physics, Engineering and Computer Science}
  \streetaddress{College Lane}
  \city{Hatfield}
  \country{United Kingdom}
  \postcode{AL10 9AB}
}

\author{Farshid Amirabdollahian}
\email{f.amirabdollahian2@herts.ac.uk}
\orcid{0000-0001-7007-2227}
\affiliation{%
  \institution{University of Hertfordshire}
  \department{School of Physics, Engineering and Computer Science}
  \streetaddress{College Lane}
  \city{Hatfield}
  \country{United Kingdom}
  \postcode{AL10 9AB}
}

\begin{abstract}
  This paper examines some common problems in Human-Robot Interaction (HRI) causing failures and troubles in Chat. A given use case’s design decisions start with the suitable robot, the suitable chatting model, identifying common problems that cause failures, identifying potential solutions, and planning continuous improvement. In conclusion, it is recommended to use a closed-loop control algorithm that guides the use of trained Artificial Intelligence (AI) pre-trained models and provides vocabulary filtering, re-train batched models on new datasets, learn online from data streams, and/or use reinforcement learning models to self-update the trained models and reduce errors.
\end{abstract}




\maketitle

\section{Introduction}

There are many scopes from which a chat can fail between humans, particularly sociolinguistic factors. Since humans are the developers of Human-Robot Interaction (HRI) chats, they can develop chat systems with the same inherent common sociolinguistic failures.  In this manuscript, the word "chat" will refer mainly to text-based chat ignoring the latency and errors caused by the spoken word recognition model if used. The author of \cite{Tannen1998argument} identifies some factors that contribute to a successful conversation between humans as follows:

\paragraph{1. Active listening:} Paying attention to what the other person is saying and actively engaging in the conversation by asking questions and showing interest. In HRI, this can be implemented by including the meaning of every word and not identifying some content to be of higher weight than others. Recent Large Language Models (LLMs) based on the transformer architecture and the attention mechanism provide this more extended sequence dependence between words in complete sentences and the context of a given corpus. This enables HRI to be more successful than human interactions because of the higher attention and memory capacity than humans, who can be distracted by a word of higher weight than another based on their cultural background or current emotional status. This statement is based on the experiences of the authors, such as the tolerance of GPS speech directions re-instructing users on rerouting options no matter how many mistakes they made. A human guide is not expected to have this infinite tolerance to errors and the ability to repeat and listen for as long as required.
\paragraph{2. Nonverbal communication:} Using body language such as eye contact, gestures, and facial expressions to reinforce what is being said. Many HRI Software Development Kits (SDKs), such as the NAOqi SDK, have animated speech models in which some body movements and facial expressions are associated with words.
\paragraph{3. Respect for differences:} Respect the other person's beliefs and opinions even if they differ from your own. Unfortunately, AI models behind robotic chats are trained on the entire internet content, encyclopedias, and huge corpus that might be biased and does not provide opinions that are inclusive of minorities or inter-racial or intercultural conversations.
\paragraph{4. Clarity of speech:} Speaking clearly and concisely to avoid misunderstandings. NAOqi SDK has default Text-to-Speech (TTS) that might be faster than normal human language and not as clear as Adults speaking the proper language accent of the given culture. There are many providers of voices of different languages, accents and gender that can be used instead to provide a human-like interaction.
\paragraph{5. Ability to stay on topic:} Staying focused on the topic at hand and avoiding tangents. Most AI chat models, such as ChatGPT, are already successful in providing answers in the exact scope of the question, as long as no ambiguous language is detected.
\paragraph{6. Equality:} Both parties have equal opportunities to express their opinions and ideas. This is spontaneously regulated in HRI, which is usually regulated on prompt/response pairs. The robot is good at waiting for the complete prompt to be finished to give a valid response based on its programmed model.

\paragraph{7. Mutual understanding:} Both parties come to a shared understanding of the conversation's purpose and goals. Most HRI chat models measure the goal of the response by relevance to the prompt they received, based on their trained dataset or feedback from the user. In some instances, ChatGPT repeated the same response several times, although the prompt it received back that its response was incorrect. There are batch training approaches vs online training. In batch training, specific organisations provide the training dataset as regulated by the laws; the dataset should not violate any laws and should not be intentionally biased. Once the training is finished, the model is used to generate responses to prompts from users, but not to be further online trained from the interaction with them. 
\paragraph{8. Open-mindedness:} Being open to new ideas and perspectives and willing to consider different viewpoints. ChatGPT usually replies with sentences such as “As an AI language model, I cannot provide opinions on my own”. As mentioned in the previous point, online learning, such as avoiding showing this batch training limitation, can enable these models to get along with a conversation and learn from it actively. This can be accomplished using a reinforcement learning algorithm. However, this might come with the dangers of providing autonomy to these models and the ability to learn from user interactions that might be dangerous overall.
\paragraph{9. Emotional intelligence:} Being aware of and managing your emotions effectively and being considerate of the other person's emotions. Some AI models detect the emotion or the sentiment of a given text, audio of the voice, or facial expression. There are even models to detect sarcasm, jokes, ambiguous sentences, and so forth. Integrating all these models with the chat AI model on open question answering might not be already implemented in any of the existing robotic assistants, but it can be recommended in future developments.
\paragraph{10. Empathy:} Understanding the other person's point of view and showing compassion. Similar to emotional intelligence, HRI AI models might not be able to be trained on all multi-modals to consider what might not be included in its training datasets to understand all cultures' points of view.

HRI chatting can be based on normal Human-Computer Interfaces (HCI) using text-based chats or speech-based chats, commonly referred to as dialogues. Personal Assistants such as Siri, Alexa and others use speech-based chats, which is an added layer of audio speech processing to identify spoken words to send to the text chat model underneath using Speech-to-Text (STT) libraries. STT can be designed using various models that can be fine-tuned to specific users. The reverse process of Text-to-Speech (TTS) is used to respond back using synthetic computerised voices that can be selected from a library. The STT/TTS layer comes with its own latency and possible errors. Historically, HRI chat was enabled using hard-coded rules, using symbolic programming, and then various text encodings of complete languages provided sub-symbolic neural architectures’ models to enhance the conversation. Robots using pattern-matching prompt/response pairs suffer more problems than those using LLMs to respond to queries. Recent advances in natural language processing (NLP) and LLMs enabled open-domain question answering, potentially enabling robots to pass the Turing test. However, common problems occur that identify the robot as a machine and not a human. This study attempts to address problems that cause chat failures with robots in different scenarios. The first section provides a non-exhaustive list of common chat problems in HRI. This is followed by a section on examples of famous failures that embarrassed the developers of these models. A conclusion of what needs to be considered in future development is provided.

\section{Common Problems}

Researching various publications and news for common problems that can arise in human-robot chat interactions, the following have been identified:

\paragraph{1. Limited conversational ability} Robots historically were programmed to respond to specific phrases or questions, and some models still follow this paradigm. This chat model does not have the ability to hold a natural conversation. These prompt/response patterns are often encoded in Artificial Intelligence Markup Language (AIML), JavaScript Object Notation (JSON), or dictionary data structures. When robots receive prompts that do not match any of the given patterns, the robot usually gives a generic response that it does not understand the human.

\paragraph{2. Natural Language Generation (NLG)} Recent advances moved the template-based prompt/response pairs into an automatically generated text using specific language symbolic understanding with hard-coded rule-based systems. This is divided into sub-tasks: content determination, text structuring, sentence aggregation, lexicalisation, referring expression generation, and linguistic realisation, which add to the system's complexity and increase its vulnerability to errors. Open-source text generation libraries exist, such as SimpleNLG and OpenCCG, which are flexible and cross-lingual, as identified in  \cite{Foster2019Natural}.

\paragraph{3. Large Language Models (LLMs)} Further advancements enabled Robots now to be programmed with access to LLMs such as ChatGPT by OpenAI or other providers such as Google, Microsoft, AWS, Nvidia, or others. These LLMs are data-driven sub-symbolic end-to-end systems using natural language embeddings that are context-aware and provide numeric vectors language representations that are closer for synonyms and distant for opposite meanings. Embeddings are trained on various corpus for many languages, providing meaningful translation with accurate performance. LLMs are pre-trained on a vast corpus that might be biased and/or limited to specific use cases in which it performs better than others \cite{Yan2022Deep}.  

\paragraph{4. Misunderstandings} These can occur when the robot can not understand what the human is saying, or the human may not understand the robot’s responses because the language model can potentially misunderstand a chat due to a variety of reasons, such as:

\begin{enumerate}[label=(\alph*)]
    \item Ambiguity in language: Human language is often complex and ambiguous, with words and phrases that can have multiple meanings. This can sometimes confuse ChatGPT or any LLM, resulting in a misinterpretation of the chat.
    \item Lack of context: Without proper context, ChatGPT or any LLM may struggle to understand the meaning behind certain words or phrases, leading to incorrect responses.
    \item Sarcasm or irony: Sarcasm or irony are often used in human communication but can be difficult for natural language processing technology to understand, resulting in a misinterpretation of the chat.
    \item Errors in input: If the chat input contains spelling mistakes, grammatical errors, or unusual sentence structures, ChatGPT or any OpenAI LLM may misconstrue the intended meaning. Although Text to Speech (TTS) and Speech To Text (STT) libraries are now used to avoid typing in the chat and provide a humanoid experience, STT models vary in accuracy, with the best being 84\% accurate and in their responsiveness \cite{Acharya20226}.
    \item Bias in training data: OpenAI LLM models are designed to learn from large volumes of data, and if that data is biased or skewed in some way, it can lead to inaccurate responses or misunderstandings in certain contexts.
\end{enumerate}

\paragraph{5. Lack of emotional intelligence} As mentioned in the introduction section, Robots or LLMs may not be able to understand or respond to human emotions in the same way as humans \cite{Shawar2007Chatbots}.

\paragraph{6. Reasoning} Mathematical, formal proofs or logical reasoning are still achieving accuracy in the range of 40:50\% approximately in fine-tuned models. If the chat context is math education, even as primary as the year three school UK curriculum, it requires logical reasoning that is still not yet available in many LLMs. Various chats require logical reasoning, not only mathematics questions/answering. Personal assistants can help meet scheduling across time differences, compare prices, or match requirements for decision-making on various projects. A fine-tuned GPT-3 model on mathematical proofs and problem answers is provided as GPT-f by the work in \cite{Polu2020Generative}. 

\paragraph{7. Technical glitches} Technical problems can disrupt the chat, such as the robot freezing, crashing, or experiencing connectivity issues with an LLM hosted in the cloud. Responsiveness is another measure of failure, such as in STT libraries that have reported less than a 300-millisecond lag in real-time transcription. Similar responsiveness is expected from a chat. The predict function call on any pre-trained model might take much longer if the model’s parameters are large or if it is hosted on the cloud with a query and prompt travel time might undergo various network connectivity problems.

\paragraph{8. Personality or cultural differences} Different cultures or personalities may have unique ways of communicating that robots may not understand, leading to misunderstandings.  This is commonly addressed by personalisation and perception training as identified in \cite{Li2022Personalized}. Social failures as well are studied by defining social failure mode and effect analysis (SFMEA) to analyse different failures, their causes, and effects. The work in \cite{Spreafico2023Artificial} applied SFMEA on Chatbots such as ChatGPT with use cases using terms from ontologies rather than generic terms. These ontologies can be design ontologies such as (e.g., function–behaviour–structure (FBS) theory), or Social sustainability ontologies such as the European Union (EU) Social Taxonomy that details the vocabulary that specifies the repercussions of the failures of social sustainability.

\paragraph{9. Lack of trustworthiness} Humans may not trust robots completely or feel uncomfortable sharing personal information with a machine. This might provide less than the required context from which the robot or an LLM can respond more accurately. This trustworthiness decreases with repeated failures or inappropriate responses \cite{Kluber2022When}.

\paragraph{10. Privacy concerns} The human may be worried about their conversations being monitored or recorded by the robot or its parent company \cite{Lewis2018Role}. Cyber security issues as well might be factors in securing privacy with Robots that are connected online.

\section{Famous Failure Examples}

The following are some examples of troubles and failures in conversations between humans and robots:

\paragraph{1. Microsoft Tay} In 2016, Microsoft launched an AI robot named Tay on Twitter, which was designed to learn from its interactions with users. However, within 24 hours, the robot began spewing out racist and sexist tweets, apparently having been corrupted by trolls and online extremists \cite{SCHWARTZ2019In}.

\paragraph{2. Amazon Alexa} Amazon’s virtual assistant Alexa has faced criticism for not understanding certain accents or dialects, causing problems for users who speak English as a second language or have a strong regional accent. In addition, Alexa’s voice recognition technology has been known to misinterpret commands, leading to user misunderstandings and frustration \cite{Lin2018Amazon's}.

\paragraph{3. Apple Siri} Apple’s virtual assistant Siri has also faced criticism for not always understanding user commands or providing accurate responses. In addition, some users have reported privacy concerns regarding Siri’s use of personal data and recordings of voice commands \cite{Hern2021Apple}.

\paragraph{4. Google Duplex} Google’s AI-powered voice assistant Duplex made headlines in 2018 for its ability to make phone calls and book appointments on behalf of users. However, some critics raised ethical concerns about the potential for the technology to deceive human call recipients by appearing to be human rather than a machine \cite{Hern2018Google}.

\paragraph{5. Sophia the Robot} Sophia is a humanoid robot developed by Hong Kong-based Hanson Robotics, which has made headlines for its realistic human-like appearance and ability to hold conversations with humans. However, some experts have criticised Sofia’s limited abilities and argued that its conversations are scripted and pre-programmed rather than truly interactive \cite{Vincent2018Facebook}.

\paragraph{6. ChatGPT} ChatGPT is an HCI that its APIs are often used now with speech layers to enable HRI conversations. It is based on the GPT-3 20 billion parameters' model that was trained using the content of the freely accessible internet. The internet is full of articles that domain experts do not validate. Some articles and links get removed, whether for disputes about their validity or re-organisation of the hosting website. The authors tested ChatGPT LLM for various contexts, such as mathematical questions, and computer science literature review, it repeatedly gave incorrect mathematical answers, even in as simple as year three math as depicted in the following example: 

\begin{verbatim}
User:
> How many even numbers between 9 and 21? Can you list them?

ChatGPT:
> There is only one even number between 9 and 21 and that is 10.
\end{verbatim}

ChatGPT has also repeatedly responded with correctly formatted but non-existent references, such as:

\noindent\texttt{Shin, J., Liu, Y., \& Oh, S. J. (2021). Retrieval\-augmented generation for knowledge-intensive question answering. arXiv preprint arXiv:2106.09659.}

\section{Conclusion}

These common problems and failure examples  illustrate some of the challenges and limitations of conversational AI technologies, which continue to evolve as researchers work to improve their accuracy, efficiency, and ethical implications. 

After determining the HRI’s use case and selecting the appropriate robot, the functional analysis can identify the areas in which the robot needs to be trained. Models, such as pre-trained models, can be downloaded for prediction upon receiving a signal from the chat STT or sensor readings. It is possible to use single-mode or multi-modal models based on the requirements. In closed-loop control algorithms, the controller can automatically choose which model to use based on the signal it receives.  These control algorithms can add a layer of ontological vocabulary selection to avoid many known social failures.

HRI is a continuous process in which a human can tell the robot that it is not responding correctly or appropriately in the given context. In order to address many problems, continuous model fine-tuning can be used when robots realise they are making mistakes. This can occur by keeping past dialogues in a local dataset to learn from for a personalised chat. The robot can also be programmed to fetch more datasets and re-training their batched models. Also, online training using data streams from sensors, IoT devices, or even the many online data streams available can solve many cognitive problems requiring multi-modal data fusing to provide human-like responses. Another possibility is to use reinforcement learning (RL) algorithms, which are more suitable for dynamic environments. RL defines the highest reward action by a utility function that maps the current state to the sequence of states leading to the highest accumulative reward, as defined for every application or environment. It starts with trial and error to learn the environment's dynamic rewards and states that are defined as a Markov Decision Process (MDP) in which new states are probabilistically defined from the current state. RL operate in two modes, exploration and exploitation. It builds its tables of state-action pairs in exploration mode to define the environmental rewards. In the exploitation model, it fetches the highest rewarding action from its tables based on learned experiences. Alternating these two modes will keep the RL agent or Robot aware of its dynamically changing environment. Various implementation details can be considered to avoid common chat failure scenarios with negative rewards from feedback from the user and encourage successful conversations by identifying the required success criteria and scores of rewards, such as integrating with a social ontology.

\bibliographystyle{wtfpaper}
\bibliography{wtfpaper}

\end{document}